\documentclass[runningheads]{llncs}
\usepackage[T1]{fontenc}
\usepackage{graphicx,verbatim}
\usepackage{subcaption}
\usepackage{amsfonts}
\usepackage{marvosym}
\usepackage{xurl}
\usepackage{hyperref}
\usepackage{color}

\begin{document}
	\title{ROTE: Routing–Timescale Framework for Balanced Expert Allocation in MoE-Based Continual Segmentation}
	\titlerunning{ROTE: Routing–Timescale Decoupling }
	
	\author{
		Zhanshi Zhu\inst{1} \and
		Shuonan Qiang\inst{1} \and
		Kuanquan Wang\inst{1}(\Letter) \and
		Shuo Li\inst{2}
		}

	%index{Zhu, Zhanshi}
	%index{Qiang, Shuonan}
	%index{Wang, Kuanquan}
	%index{Li, Shuo}

	\authorrunning{Z. Zhu et al.}

	\institute{
		Faculty of Computing, Harbin Institute of Technology,
		Harbin, China
		\email{wangkq@hit.edu.cn}
		\and
		Department of Computer and Data Science and
		Department of Biomedical Engineering,
		Case Western Reserve University,
		Cleveland, USA
	}
	
	\maketitle              % typeset the header of the contribution

	\begin{abstract}
		Continual segmentation is essential in dynamic clinical environments, where models must adapt to evolving imaging protocols and anatomical targets without forgetting previously learned knowledge. In continual settings, embedding Mixture-of-Experts (MoE) modules into the Segment Anything Model (SAM) decoder enables scalable capacity expansion without fine-tuning the backbone.
		However, existing MoE gating mechanisms are often challenged by expert allocation imbalance, where routing weights rapidly collapse onto a small subset of experts. This results in under-utilized capacity, degraded expert reusability, and aggravated catastrophic forgetting over long task sequences.
		To address challenges of expert allocation imbalance and catastrophic forgetting, we propose ROTE, a ROuting–TimEscale framework for continual segmentation with MoE-augmented SAM. ROTE integrates: (1) Locally-Regularized Soft Routing with KAN enforces balanced routing to prevent early expert monopolization; and (2) Hierarchical Temporal Learning inspired by nested learning decouples updates across time scales between SAM backbone, routing and expert to maintain long-term continual learning.
		Extensive experiments show ROTE consistently improves segmentation performance and mitigates catastrophic forgetting.
		The implementation code is publicly available at:
		\url{https://github.com/PerceptionComputingLab/ROTE}.
	\end{abstract}
	% Authors must provide keywords and are not allowed to remove this Keyword section.
	
	%
	%
	%
	
	\section{Introduction}
	
	\begin{figure}[t]
		\centering
		\includegraphics[width=\linewidth]{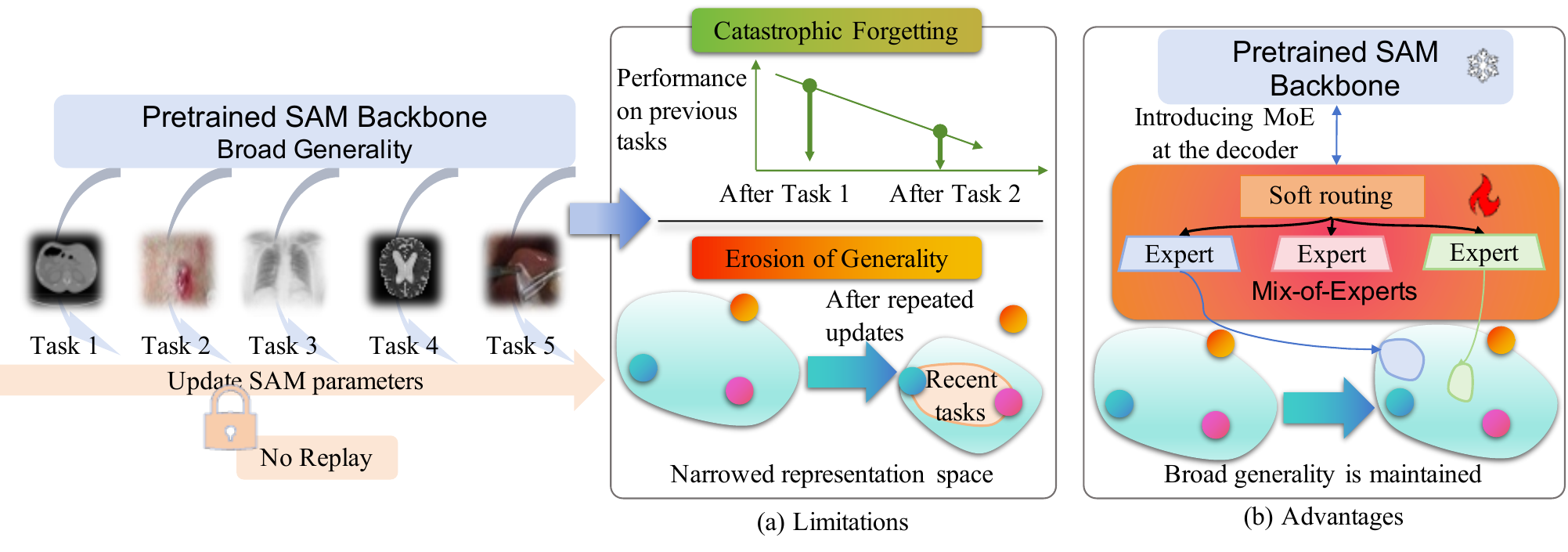}
		\caption{
			(a) SAM-based continual segmentation requires updating the SAM backbone, compromising generality.
			(b) MoE-Augmented SAM has advatantages on maintaining generality. 
		}
		\label{issues}
	\end{figure}
	
	Under the no-replay constraint, directly using Segment Anything Models (SAM) \cite{kirillov2023segmentanything} for continual segmentation suffers from limitations (Fig.~\ref{issues}(a)): eroding generality and amplifying catastrophic forgetting \cite{mccloskey1989catastrophic} due to repeatedly updates backbone parameters. 
	In real-world clinical deployment, segmentation systems continuously face distribution shifts (e.g., new centers, scanner changes, and protocol variations), while historical data are often unavailable due to privacy or storage constraints \cite{bruno2025continual}. SAMs provide an appealing unified backbone for medical segmentation thanks to strong general-purpose representations and a prompt-driven	interface \cite{huang2024segment,ma2024medsam}. However, without expandable modular capacity, continual adaptation is forced to keep perturbing shared parameters, causing accumulated forgetting and degraded cross-task generalization. Replay-based continual learning can mitigate forgetting, but its reliance on storing and accessing historical data often conflicts with clinical data governance requirements.
	
	SAM augmented by Mixture-of-Experts (MoE) \cite{shazeer2017moe} enables modular capacity expansion and decouples plasticity from the shared backbone, making scalable continual learning possible and maintained broad generality (Fig.~\ref{issues}(b)).
	MoE provides conditional capacity by routing each input to a mixture of specialized experts, so new domains and tasks can be absorbed by experts while the shared backbone remains stable—an especially natural fit
	for foundation-model-centric continual segmentation \cite{Wan_SAMMed3DMoE_MICCAI2024,zhang2025intelligent}. Introducing MoE at the decoder stage injects additional adaptation capacity without sacrificing SAM's generality, supporting heterogeneous
	distributions across centers and time. Crucially, in continual learning with a frozen foundation model, gradient accessibility and expert recoverability are often more important than sparse computation efficiency; thus, a soft (dense) routing formulation is preferable for flexible expert composition and stable long-horizon reuse \cite{shazeer2017moe,zhou2022expertchoice,plotka2025mamba}.
	
\begin{figure}[t]
		\centering
		\includegraphics[width=\linewidth]{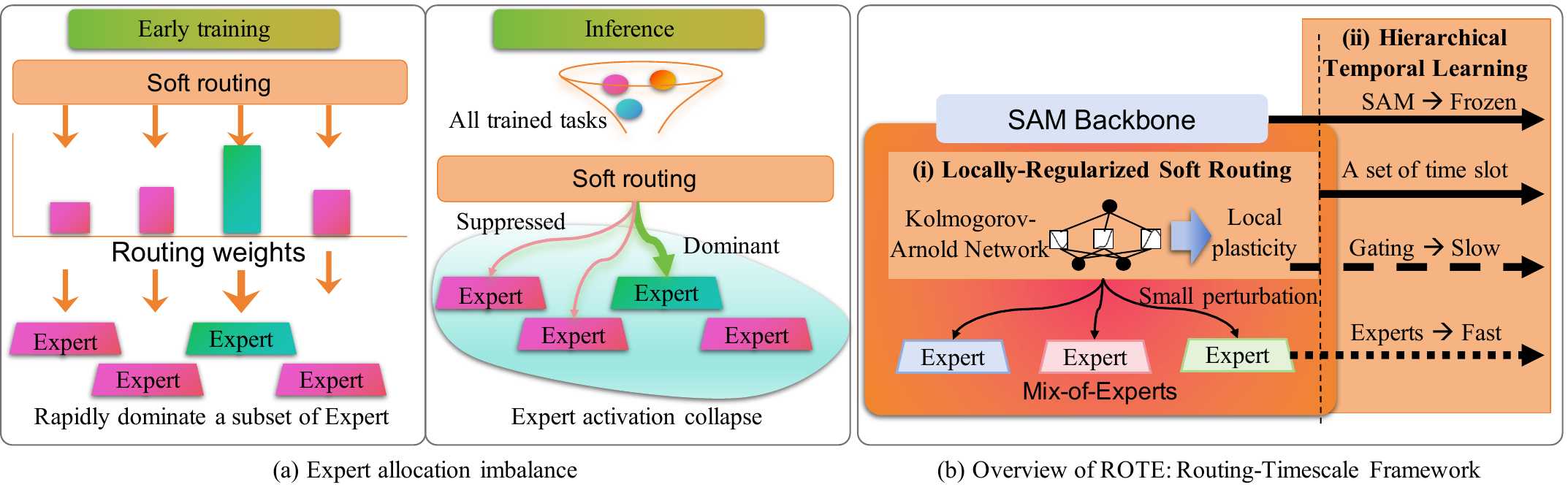}
		\caption{
			(a) Expert allocation imbalance: the gate quickly concentrates routing weights on a few experts early in training, leading to expert activation collapse at inference.
			(b)	The ROTE framework integrates routing-aware and timescale-aware designs to promote balanced expert utilization and
			long-term continual learning.
		}
		\label{methods}
	\end{figure}

	However, MoE-augmented SAM exhibits a critical challenge beyond catastrophic forgetting: expert allocation imbalance, which drives expert activation collapse and undermines capacity reuse  (Fig.~\ref{methods}(a)). Early in training, standard gating mechanisms tend to rapidly concentrate routing weights on a small subset of experts, resulting in highly uneven gradient updates and leaving most experts under-trained \cite{shazeer2017moe,zhou2022expertchoice}. As the task stream grows, this imbalance compounds: a few experts become monopolized and prematurely tied to early tasks, suppressing expert reuse and recomposition for later shifts and reducing overall plasticity and transferability \cite{mallya2018packnet}. At inference, the issue manifests as expert activation collapse, where only a handful of experts are consistently activated while the rest contribute negligibly. Over long task sequences, such routing collapse progressively nullifies the reusable capacity promised by soft MoE, limiting the long-term gains of routing-based continual segmentation.
	
	We propose \textbf{ROTE} (\textbf{Ro}uting-\textbf{T}im\textbf{E}scale), a framework that jointly allivates expert allocation imbalance and mitigates forgetting by combining routing regulation with timescale decoupling for long-term continual segmentation.
	ROTE keeps the SAM backbone frozen to maximize the stability of general-purpose representations, while introducing MoE experts	exclusively in the decoder and adopting continuously weighted soft routing to preserve gradient accessibility and expert recoverability (Fig.~\ref{methods} (b)). (i) From the \textbf{routing perspective}, we propose Locally-Regularized Soft Routing (LRSR) based on Kolmogorov--Arnold Networks (KAN) \cite{liu2025kan}, which regulates routing behavior via constrained nonlinear mappings \cite{wang2025medkaformer}. Its locally plastic mapping delays early routing collapse, promotes balanced gradient allocation, and preserves multi-expert activation at inference to improve capacity utilization. (ii) From the \textbf{timescale perspective}, we propose Hierarchical Temporal Learning (HTL), inspired by the timescale separation principle of nested learning (NL) \cite{behrouz2025nested}, to decouple component update dynamics \cite{jafari2025dynamic}: the backbone remains static, decoder experts are updated rapidly to adapt to incoming tasks, and routing functions evolve slowly to maintain	cross-task consistency. This timescale decoupling establishes a controllable stability-plasticity balance, reducing forgetting
	and improving long-term continual segmentation.
	
	In summary, the main contributions of this work are threefold:
	\begin{itemize}
		\item We identify and formalize expert allocation imbalance as a challenge of soft MoE-augmented,
		frozen-foundation continual segmentation, and are the first to address it via \textbf{ROTE}---a routing-timescale framework.
		\item We propose Locally-Regularized Soft Routing (LRSR) based on KAN to ensure smooth and balanced expert routing through constrained nonlinear functions, enabling effective expert reuse over long task streams.
		\item We propose Hierarchical Temporal Learning (HTL) inspired by NL to decouple the update dynamics of model components across time scales, improving the stability-plasticity trade-off and long-term continual learning.
	\end{itemize}
	
	\section{Method}
	ROTE integrates two key components: (1) Locally-Regularized Soft Routing (LRSR) based on Kolmogorov--Arnold Networks (KAN)  to balance routing, preventing early expert monopolization; and (2) Hierarchical Temporal Learning (HTL), inspired by Nested Learning (NL), to decouple update dynamics across time scales, matigating forgertting.

	\subsection{Problem Setup}
	
	In continual segmentation, we consider a stream of tasks
	$
	\mathcal{T} = \{ \mathcal{T}_1, \mathcal{T}_2, \dots, \mathcal{T}_T \},
	$
	where the $t$-th task $\mathcal{T}_t$ is associated with a dataset
	$
	\mathcal{D}_t = \{(p_i^{(t)}, x_i^{(t)}, y_i^{(t)})\}_{i=1}^{N_t}.
	$
	During training on $\mathcal{T}_t$, data from previous tasks $\mathcal{D}_{<t}$ is inaccessible.
	
	The segmentation model builds upon a pre-trained SAM, which performs interactive segmentation conditioned on prompt inputs $p_i^{(t)}$. To enable continual adaptation under a frozen backbone, a \textit{Soft Mixture-of-Experts} (MoE) module is embed into SAM's mask decoder. The MoE module contains $K$ experts $\{E_k(\cdot;\theta_k)\}_{k=1}^K$ and a continuous routing function:
	\[
	\mathbf{w}(\cdot;\phi)
	=
	[w_1(\cdot;\phi), \dots, w_K(\cdot;\phi)],
	\quad
	w_k(\cdot;\phi)\ge 0,\;
	\sum_{k=1}^K w_k(\cdot;\phi)=1,
	\]
	which modulates expert contributions through soft weighting during decoding.
	
	For task $\mathcal{T}_t$, the model is optimized by minimizing the segmentation loss
	$
	\mathcal{L}_t =
	\mathbb{E}_{(p,x,y)\sim\mathcal{D}_t}
	\big[
	\ell(f(x,p;\theta_{\mathrm{SAM}},\Theta,\phi),\,y)
	\big],
	$
	where $\Theta = \{\theta_k\}_{k=1}^K$ are the expert parameters, $\phi$ denotes routing parameters, and $\theta_{\mathrm{SAM}}$ is the frozen backbone. During training, only $\Theta$ and $\phi$ are updated via:
	$
	\theta_k \leftarrow \theta_k - \eta_\theta \nabla_{\theta_k}\mathcal{L}_t,
	\quad
	\phi \leftarrow \phi - \eta_\phi \nabla_{\phi}\mathcal{L}_t.
	$
	
	\subsection{Locally-Regularized Soft Routing via KAN}
	
	To mitigate premature expert dominance caused by gradient-driven weight concentration in continual learning, a LRSR based on KAN is proposed to enhance the soft routing function in MoEs. Each expert $E_k$ receives a continuous weight $w_k(\cdot;\phi)$, where $\phi$ denotes the routing parameters optimized via the segmentation loss.
	During sequential training, if the routing function is overly sensitive to inputs or parameter perturbations, its gradient
	$
	\nabla_{\phi} \mathcal{L}_t =
	\sum_{k=1}^K
	\frac{\partial \mathcal{L}_t}{\partial w_k}
	\cdot
	\frac{\partial w_k}{\partial \phi}
	$
	tends to be dominated by a small subset of experts. This leads to early routing weight collapse, where only a few experts receive updates while others remain under-utilized.
	
	LRSR parameterizes the routing function using piecewise spline functions of KAN, whose key property is local plasticity\cite{wang2025medkaformer}. Specifically, each spline parameter $\phi_j$ affects the routing function only within a localized input region. Formally, we have
	$
	\frac{\partial w_k}{\partial \phi_j} \approx 0,
	$
	if the input lies outside the support of $ \phi_j$.
	This structural constraint enforces strong locality in gradient updates, effectively suppressing indiscriminate global gradient propagation across all experts.
	
	From a gradient propagation perspective, this locality of routing yields two key effects.
	First, small perturbations in inputs or parameters no longer induce disproportionate changes in routing behavior. As a result, the gradient norm
	\(
	\|\nabla_{\phi} w_k\|
	\)
	remains stable throughout training, enhancing the robustness of expert weighting.
	Second, the gradient contributions from different experts,
	$
	\frac{\partial \mathcal{L}_t}{\partial w_k}
	\cdot
	\frac{\partial w_k}{\partial \phi},
	$
	become more comparable in magnitude. This prevents any single expert from exponentially dominating the routing updates, thereby promoting balanced expert utilization over time.
	
	\textbf{Summary of Innovation}. In the continual learning setting, the LRSR enables locally controlled plasticity for expert routing. Instead of rapidly collapsing onto a subset of experts during early training, the routing weights evolve gradually across tasks, allowing more balanced expert participation. Compared to MLP- or attention-based gating functions, which lack explicit gradient locality, KAN structurally constrains gradient propagation and suppresses premature divergence across experts. This effectively delays routing collapse and preserves expert utilization efficiency over long task sequences.
	
	\subsection{Hierarchical Temporal Learning via Nested Learning}
	
	The HTL strategy, inspired by NL, decomposes the update dynamics of expert and routing parameters through time-scale separation, aiming to further mitigate catastrophic forgetting. From a gradient propagation perspective, routing parameters $\phi$ modulate the magnitude and direction of gradients received by each expert via the weighting function $\mathbf{w}(\cdot;\phi)$.
	When routing and expert parameters are updated at the same time scale, $\phi$ tends to overfit transient task-specific gradients, leading to frequent reorganization of expert weights at task boundaries. Consequently, previously learned experts are repeatedly exposed to disruptive gradient signals from new tasks, destabilizing continual learning.
	
	HTL addresses this issue by enforcing explicit time-scale separation between fast- and slow-updating components. In MoE-augmented SAM framework, parameters naturally fall into three timescale categories: (i) the SAM backbone parameters $\theta_{\mathrm{SAM}}$, which remain frozen to preserve representational stability; (ii) the expert parameters $\Theta = \{\theta_k\}$, which are updated at every training step to adapt to current tasks; and (iii) the routing parameters $\phi$, which modulate expert usage across tasks and influence gradient flow over extended time horizons.
	To prevent short-term overfitting, routing parameters are updated less frequently than experts, while SAM parameters are frozen. Specifically, while $\theta_k$ are updated at every step, $\phi$ is updated only once every $T$ steps. Equivalently, the routing gradient can be viewed as an accumulated sum over a temporal window:
	$
	\nabla_{\phi}\mathcal{L}
	\approx
	\sum_{j=t-T+1}^{t} \nabla_{\phi} \mathcal{L}_j.
	$
	This update scheme reduces the sensitivity of routing to batch-level noise, allowing it to encode longer-term gradient trends and improve timescale stability.
	
	This timescale separation in HTL yields two important effects at the gradient level.
	First, the evolution of routing weights becomes explicitly smoothed, preventing abrupt reconfigurations during task transitions and preserving cross-task consistency in expert allocation.
	Second, under a relatively stable routing context, expert parameters can rapidly adapt to new tasks without being repeatedly disrupted by shifting gradient flows. This separation of adaptation and coordination allows experts to specialize more effectively across sequential tasks.
	
	\textbf{Summary of Innovation}. The HTL constitutes a hierarchical optimization system composed of fast variables (expert parameters), slow variables (routing parameters), and frozen vatiables (SAM backbone parameters). The slowly evolving routing parameters provide a stable context for gradient allocation, enabling fast-updating experts to adapt efficiently to new tasks. In turn, expert performance over multiple steps guides the long-term evolution of the routing function. Meanwhile, the SAM backbone preserves representational stability. Through this nested decoupling of time scales, the model achieves a principled balance between plasticity and stability, maintaining adaptability to new tasks while effectively mitigating catastrophic forgetting.
	
	\section{Experiments}
	
	\subsection{Experimental Settings}
	
	\textbf{Dataset and task setup.}
	ROTE is evaluated on a benchmark comprising 10 publicly available medical image segmentation datasets, spanning a wide range of imaging modalities including endoscopy, fundus photography, X-ray, CT, MRI, OCT, and histopathology. Each dataset involves distinct anatomical structures and segmentation targets. These datasets are arranged into a sequential task stream, where each task introduces a previously unseen modality or anatomical region, resulting in substantial cross-task distribution shifts. All tasks adopt a unified train/validation/test split with a ratio of 7:1:2.
	
	\textbf{Implementation details.}
	All experiments are conducted using the SAM, initialized with pretrained weights from the COSMOS-1050K dataset\cite{huang2024segment}. During continual learning, the SAM backbone remains entirely frozen. The decoder is extended with a Soft MoE module, and only the parameters of the MoE are updated. The Adam optimizer is used with an initial learning rate of $1 \times 10^{-4}$. Tasks are learned sequentially without access to data from previous tasks, following standard continual learning protocols.
	
	\textbf{Evaluation metrics.}
	Segmentation performance is assessed using the Dice Similarity Coefficient (DSC). To characterize continual learning behavior, we additionally report Incremental Learning DSC (IL-DSC), which evaluates segmentation performance accumulated over all seen tasks, and Backward Transfer (BWT), which quantifies the retention of knowledge on previously learned tasks \cite{li2022domain}. All metrics are computed on the held-out test sets.
	\begin{table}[t]
		\centering
		\caption{
			Our ROTE achieves the best performance in terms of both overall segmentation accuracy and degrading forgetting, mesured by 
			IL-DSC and BWT,	respectively. Higher values indicate better performance.
		}
		\label{tab:overall}
		\begin{tabular}{lcc}
			\hline
			Method & IL-DSC $\uparrow$ & BWT $\uparrow$ \\
			\hline
			Fine-tuning (Decoder-only adaptation) 
			& 90.29 & 96.61 \\
			MR \cite{bera2023memory} (Replay-based with memory buffer)        
			& 90.48 & 96.87 \\
			LwF \cite{kirkpatrick2017overcoming} (Output-level distillation)   
			& 90.35 & 96.68 \\
			Refresh \cite{wang2024unified} (Regularization-constrained updates) 
			& 90.21 & 96.48 \\
			CODA-Prompt \cite{smith2023coda} (Prompt-based task modulation) 
			& 89.38 & 98.71 \\
			\hline
			Soft MoE (MLP gating)        
			& 90.61 & 97.10 \\
			Soft MoE + LRSR (KAN; Structured nonlinear routing)             
			& 90.61 & 97.13 \\
			Soft MoE + HTL (NL; Time-scale separated optimization)      
			& 90.67 & 97.12 \\
			\textbf{ROTE (LRSR + HTL)}    
			& \textbf{91.52} & \textbf{97.70} \\
			\hline
		\end{tabular}
	\end{table}

	\subsection{Comparison with Other Methods}
	
	Table~\ref{tab:overall} demonstrates that the our ROTE achieves the best performance in terms of both segmentation accuracy and forgetting suppression, validating its effectiveness for continual medical image segmentation. First, under a unified experimental setting, ROTE attains the highest scores on both IL-DSC and BWT, indicating that it can maintain strong overall segmentation performance while preserving previously learned knowledge as tasks accumulate. In contrast, other continual learning approaches exhibit noticeable performance degradation under severe cross-modality and cross-target distribution shifts: decoder-only fine-tuning lacks explicit mechanisms for cross-task stability, while replay-, distillation-, and regularization-based methods rely on historical samples or additional constraints and still struggle to balance accuracy and knowledge retention over long task sequences. Furthermore, Soft MoE with a standard MLP gate, as a direct capacity expansion baseline, achieves only 90.61 IL-DSC and 97.10 BWT, which is clearly inferior to the proposed approach. This result suggests that simply introducing an MoE structure is insufficient for continual learning.
	
	\subsection{Ablation Study}
	
	The combination of LRSR and HTL achieves both high accuracy and stable knowledge retention.
	Table~\ref{tab:overall} summarizes the overall continual segmentation performance under different architectural variants.
	Starting from a Soft MoE baseline with conventional MLP gating, incorporating either LRSR or HTL alone leads to consistent but limited improvements.
	Specifically, Soft MoE + LRSR primarily improves BWT by stabilizing expert allocation and enhancing expert reuse, while Soft MoE + HTL yields moderate gains in both IL-DSC and BWT by improving timescale stability.
	Notably, integrating both components in ROTE results in a substantial performance boost, achieving the best IL-DSC and BWT among all compared methods.
	These results indicate that routing regularization and timescale decoupling should be jointly considered to effectively mitigate forgetting.

\begin{figure}[t]
		\centering
		\includegraphics[width=0.70\linewidth]{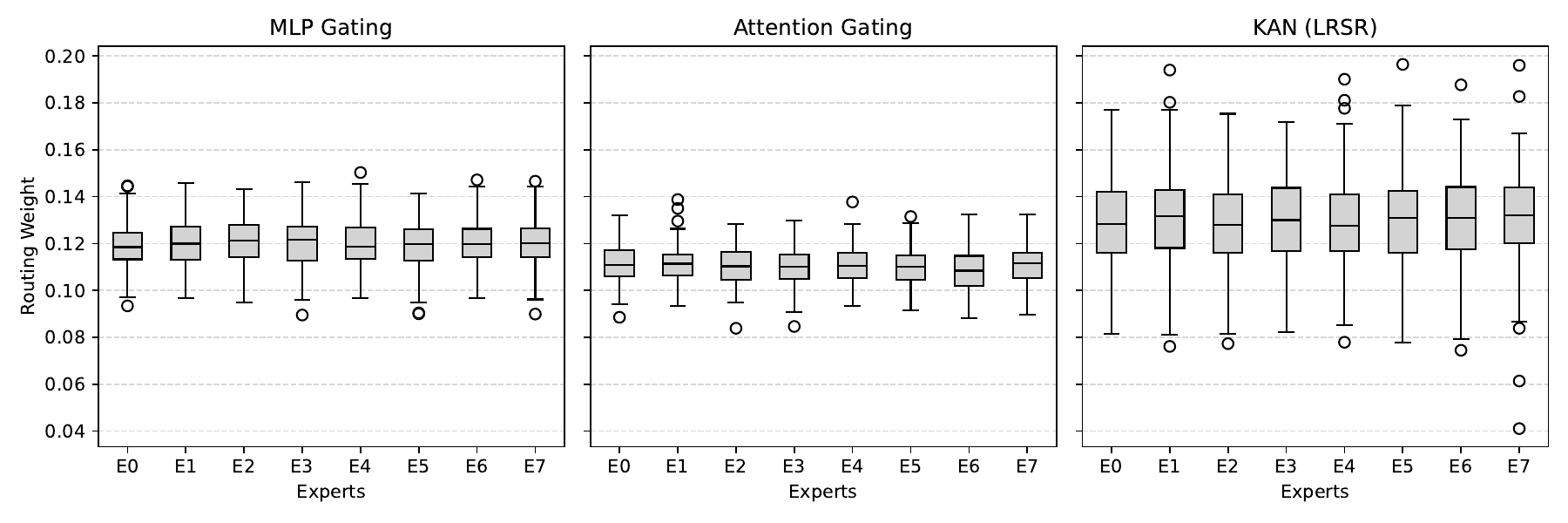}
		\hfill
		\includegraphics[width=0.29\linewidth]{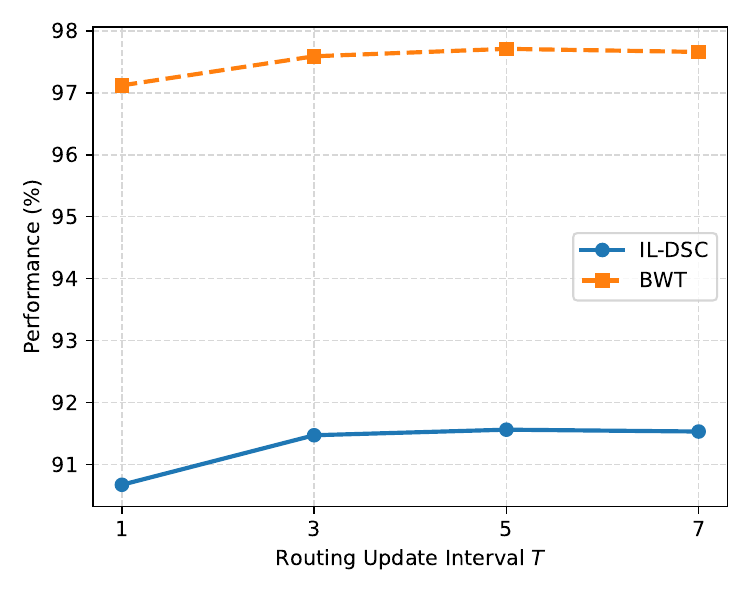}
		\caption{
			Effectiveness analysis of routing- and timescale-level designs in ROTE.
			\textbf{(Left)} Distribution of expert routing weights under different gating mechanisms.
			LRSR preserves broader and more balanced expert participation, mitigating early expert allocation collapse.
			\textbf{(Right)} Impact of HTL under different routing update intervals $T$.
			$T=1$ corresponds to the absence of HTL.
			Introducing time-scale separation ($T \geq 2$) substantially improves both IL-DSC and BWT, yielding stable performance across a wide range of update intervals.
		}
		\label{fig:ablation_routing_temporal}
	\end{figure}

	LRSR alleviates expert allocation imbalance.
	Without routing regularization, MoE gating rapidly collapses onto a small subset of experts during continual training, leading to severe expert allocation imbalance.
	Fig.~\ref{fig:ablation_routing_temporal} (left) visualizes the distribution of expert routing weights under different gating mechanisms in reference.
	Both MLP- and attention-based gating exhibit highly concentrated weight distributions, indicating premature expert monopolization.
	In contrast, the proposed KAN-based LRSR maintains broader and more evenly distributed routing weights across experts throughout training.
	This behavior demonstrates that LRSR structurally suppresses early routing collapse by enforcing localized gradient propagation, thereby preserving expert plasticity and sustained utilization over long task sequences.
	
	Time-scale separation in HTL is critical for mitigating forgetting.
	Updating routing and expert parameters at the same time scale causes unstable expert coordination and exacerbates catastrophic forgetting.
	Fig.~\ref{fig:ablation_routing_temporal} (right) examines the impact of HTL by varying the routing update interval $T$.
	When routing parameters are updated at the same frequency as expert parameters ($T=1$), both IL-DSC and BWT degrade noticeably.
	Introducing HTL ($T \geq 2$) substantially improves performance, with consistent gains in both metrics. Moreover, performance remains stable across a wide range of update intervals, indicating that HTL does not rely on sensitive hyperparameter tuning. These results confirm that decoupling routing and expert updates across time scales is essential for stabilizing expert coordination and enabling long-term continual segmentation.
	
	\section{Conclusion}
	
	In this work, we propose ROTE, a routing–timescale framework for continual medical image segmentation built upon a frozen SAM backbone with MoE-augmented decoders.
	To address expert allocation imbalance and catastrophic forgetting, ROTE integrates LRSR to balance expert allocation and HTL to decouple backbone, routing and expert updates across time scales for mitigating catastrophic forgetting.
	Extensive experiments demonstrate that ROTE achieves superior overall segmentation performance over long task sequences.

	\begin{credits}
		\subsubsection{\ackname} This study was funded
		by the National Natural Science	Foundation of China under Grants 62272135 and 82527807.
		
		\subsubsection{\discintname}
		The authors have no competing interests to declare that are
		relevant to the content of this article. 
	\end{credits}
	
	\bibliographystyle{splncs04}
	\bibliography{mybibliography}

\end{document}